\documentclass[10pt, a4paper]{article}
\usepackage{lrec}
\usepackage{multibib}
\newcites{languageresource}{Language Resources}
\usepackage{graphicx}
\usepackage{tabularx}
\usepackage{soul}
\usepackage{arabtex}
\usepackage{booktabs}
\usepackage{utf8}
\usepackage{kotex}
\usepackage{arydshln}
\usepackage[noend]{algpseudocode}
\usepackage[utf8]{inputenc}
\usepackage[table,xcdraw]{xcolor}
\usepackage{epstopdf}
\usepackage[utf8]{inputenc}
\usepackage{mdframed}
\usepackage{hyperref}
\usepackage{xstring}
\usepackage{enumitem}
\usepackage{color}
\usepackage{mathtools}
\usepackage{hyperref}
\hypersetup{
    colorlinks=true,
    linkcolor=blue,
    citecolor=blue,
    filecolor=magenta,      
    urlcolor=cyan,
}
\usepackage{adjustbox}
\usepackage{booktabs}
\usepackage{multirow}
\usepackage{latexsym}
\usepackage{arabtex}
\usepackage{comment}
\usepackage{array}

\newcolumntype{H}{>{\setbox0=\hbox\bgroup}c<{\egroup}@{}}

\DeclareFixedFont{\ttb}{T1}{txtt}{bx}{n}{12} 
\DeclareFixedFont{\ttm}{T1}{txtt}{m}{n}{12}  

\usepackage{color}
\definecolor{deepblue}{rgb}{0,0,0.5}
\definecolor{deepred}{rgb}{0.6,0,0}
\definecolor{deepgreen}{rgb}{0,0.5,0}

\usepackage{listings}

\newcommand\pythonstyle{\lstset{
language=Python,
basicstyle=\ttm,
morekeywords={self},              
keywordstyle=\ttb\color{deepblue},
emph={MyClass,__init__},          
emphstyle=\ttb\color{deepred},    
stringstyle=\color{deepgreen},
frame=tb,                         
showstringspaces=false
}}

\lstnewenvironment{python}[1][]
{
\pythonstyle
\lstset{#1}
}
{}

\newcommand\pythoninline[1]{{\pythonstyle\lstinline!#1!}}

\usepackage{listings}
\usepackage{color}

\definecolor{dkgreen}{rgb}{0,0.6,0}
\definecolor{gray}{rgb}{0.5,0.5,0.5}
\definecolor{mauve}{rgb}{0.58,0,0.82}

\newmdenv[leftline=false,rightline=false,topline=false]{topbot}
\newcommand\blfootnote[1]{%
  \begingroup
  \renewcommand\thefootnote{}\footnote{#1}%
  \addtocounter{footnote}{-1}%
  \endgroup
}

\title{TURJUMAN: \\A Public Toolkit for Neural Arabic Machine Translation      \\ \vspace*{.5\baselineskip}}

\name{El Moatez Billah Nagoudi${^{\star}}$~~~ AbdelRahim Elmadany${^{\star}}$~~~Muhammad Abdul-Mageed${^{\star}}$}
\address{Deep Learning \& Natural Language Processing Group  \\
         The University of British Columbia \\
         \{moatez.nagoudi,a.elmadany,muhammad.mageed\}@ubc.ca\\}

\abstract{
We present TURJUMAN, a neural toolkit for translating from $20$ languages into Modern Standard Arabic (MSA). TURJUMAN exploits the recently-introduced text-to-text Transformer AraT5 model, endowing it with a powerful ability to decode into Arabic. The toolkit offers the possibility of employing a number of diverse decoding methods, making it suited for acquiring paraphrases for the MSA translations as an added value. To train TURJUMAN, we sample from publicly available parallel data employing a simple semantic similarity method to ensure data quality. This allows us to prepare and release AraOPUS-20, a new machine translation benchmark. We publicly release our translation toolkit (TURJUMAN) as well as our benchmark dataset (AraOPUS-20).$\textcolor{blue}{^{1}}$\\ \newline \Keywords{Machine Translation, Neural Machine Translation, Arabic, Arabic NLP, Open Source, TURJUMAN, Toolkit}. }

\begin{document}
\setcode{utf8}
\maketitleabstract

\section{Introduction}\label{sec:introduction}
Natural language processing (NLP) technologies such as question answering, machine translation (MT), summarization, and text classification are witnessing a surge. This progress is the result of advances in deep learning methods, availability of large datasets, and increasingly powerful computing infrastructure. As these technologies continue to mature, their applications in everyday life become all the more pervasive. For example, neural machine translation (NMT), the focus of the current work, has applications in education, health, tourism, search, security, recreation, etc. Similar to other areas, progress in NMT is contingent on high-quality, standardized datasets and fast prototyping. Such datasets and tools are necessary for meaningful comparisons of research outcomes, benchmarking, and training of next generation scholars. Off-the-shelf tools are also especially valuable both as stand-alone and enabling technologies in all research and development. Although various tools have been developed for Arabic NLP tasks such as those involving morphosyntactic analysis~\cite{2014-madamira,2016-farasa,obeid-etal-2020-camel} and detection of social meaning~\cite{abdul2019aranet,farha2019mazajak}, there has not been as much progress for MT. More specifically, there is shortage of publicly available tools for Arabic MT. The goal of this work is to introduce \textbf{TURJUMAN}, a new publicly available Arabic NMT toolkit that seeks to contribute to bridging this gap. ~\blfootnote{ $^{1}$\url{https://github.com/UBC-NLP/turjuman}} ~\blfootnote{ $^{\star}${All authors contributed equally.}}\\

Recent advances in NMT leverages progress in Transformer-based encoder-decoder language models, and TURJUMAN takes advantage of such a progress. In particular, encoder-decoder models such as  MASS~\cite{song2019mass},  BART~\cite{lewis2020bart}, and T5\cite{raffel2019exploring}, and their multilingual counter-parts have all been shown to remarkably benefit NMT. For this reason, TURJUMAN is built off AraT5~\cite{nagoudi2022_arat5}. AraT5 is a recently released text-to-text Transformer model. For comparisons, we benchmark AraT5 against a number of baselines on a new parallel dataset that we also introduce as part of this work. Importantly, we do not intend TURJUMAN as a tool for delivering state-of-the-art (SOTA) translations. For this reason, we do not use all parallel datasets at our disposal. Rather, we introduce TURJUMAN as an extensible framework. For example, it can be further developed to produce SOTA performance by fine-tuning its backend model on larger datasets.

\begin{figure}[t]
  \centering
  \includegraphics[width=\linewidth]{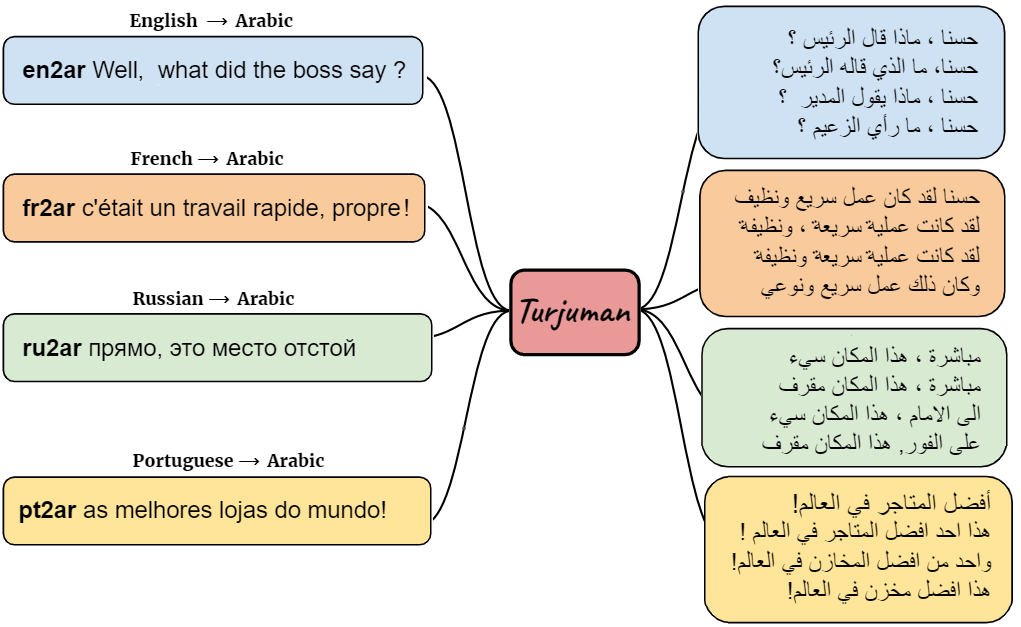}
\caption{\small Our TURJUMAN neural machine translation toolkit illustrated with four prompt MT tasks: English, French, Russian, and Portuguese $\rightarrow$ Arabic. For each source sentence, we  employ four decoding methods to produce output:  \textit{greedy search, beam search, top-k sampling}, and \textit{top-p sampling}.}
\label{fig:trj_overview} 
\end{figure}


\par In the context of creating our tool, we also prepare and release \textbf{AraOPUS-20}. AraOPUS-20 is a reasonably-sized parallel dataset of $20$ language pairs (with X $\rightarrow$ Arabic) for NMT. We extract AraOPUS-20 from OPUS~\cite{OPUS}. Since OPUS is known to involve noisy translations, we propose a simple quality assurance method based on semantic similarity to remove this noise from the dataset. We release AraOPUS-20 in standard splits, thereby making it well-suited for Arabic MT model comparisons. 

TURJUMAN also integrates recent progress in diverse decoding, such as \textit{greedy search}~\cite{cormen2009introduction}, \textit{beam search}~\cite{koehn2009statistical}, \textit{top-k sampling}~\cite{fan2018hierarchical}, and \textit{nucleus sampling}~\cite{holtzman2019curious}. This makes it possible to use TURJUMAN for generating various translations of the same foreign sequence. As such, TURJUMAN can also be used for producing paraphrases at the Arabic side (see Figure~\ref{fig:trj_overview}).



To summarize, we make the following contributions:

 \begin{enumerate}
 \item We prepare and release \textbf{\textit{AraOPUS-20}}, an MT benchmark that we extract  from the freely available parallel corpora OPUS~\cite{OPUS}. AraOPUS-20 consists of bitext between Arabic and $20$ languages. The languages paired with Arabic include high-resource languages such as English, French, and Spanish and low-resource ones such as Cebuano,\footnote{Language spoken in the southern Philippines} Tamashek,\footnote{Tamashek or Tamasheq is a variety of Tuareg, a Berber macro-language widely spoken by nomadic tribes across North Africa countries.} and Yoruba\footnote{Yoruba is a language spoken in West Africa, primarily in Southwestern Nigeria.}.

 \item We introduce \textbf{\textit{TURJUMAN}}, a python-based  NMT toolkit for translating sentences from $20$ languages into Arabic. TURJUMAN fine-tunes AraT5~\cite{nagoudi2022_arat5}, a powerful Arabic text-to-text Transformer language model. Our toolkit can be used off-the-shelf as a strong baseline, or as an enabling technology. It is also extensible. For example, it can be further developed through additional fine-tuning on larger amounts of data. 
 
  \item We endow TURJUMAN with a diverse set of decoding capabilities, making it valuable for generating paraphrases~\cite{fadaee2017data} of foreign content into Arabic. 
 

 \end{enumerate}

The rest of the paper is organized as follows: We provide an overview of works related  to Arabic machine translation in  Section~\ref{sec:RW} We introduce AraOPUS-20 MT benchmark  in Section~\ref{sec:AraOPUS-20-Data} We describe TURJUMAN in Section~\ref{sec:Torjuman}, and Section~\ref{sec:concl} is where we conclude.

\section{Related Work}\label{sec:RW}

\noindent Our work is related to research on MT datasets and tools, and language models on which these tools may be fine-tuned. Hence, we start our coverage of related work by presenting most of the popular Arabic MT datasets for both MSA and Arabic dialects. We then provide an overview of Arabic MT systems and tools.  Finally, we review both Arabic and multilingual encoder-decoder pre-trained language  models since these are most relevant to the translation task.

\subsection{MSA MT Resources}
\label{subsec:MT_Resources_MSA}
\vspace{3mm}

\textbf{Open Parallel Corpus (OPUS).}~\newcite{OPUS} propose the  large, multi-lingual, parallel  sentences datasets OPUS. OPUS contains more than $2.7$ billion parallel sentences in $90$ languages including Arabic.\footnote{\url{https://opus.nlpl.eu/}} We extract AraOPUS-20 from OPUS. A number of additional MSA  datasets involving Arabic have also been proposed. Although we do not make use of any of these, we review them here both for completeness and since they can be exploited for extending TURJUMAN.

  \begin{table*}[t]
\centering
 \renewcommand{\arraystretch}{0.7}
\resizebox{0.7\textwidth}{!}{%
\begin{tabular}{llr}
\toprule

\textbf{MSA sentences}
\scriptsize & \textbf{English sentences}                                                                 & \multicolumn{1}{l}{\textbf{Sim}} \\ \toprule
\colorbox{red!40}{\<ألمانيا>}	 & \colorbox{red!40}{The annex to the present report.} & \colorbox{red!40}{-0.12 } \\
\footnotesize  \colorbox{red!35}{\< إساءة استغلال مركز الهيمنة>}	 & \colorbox{red!35}{Abuse of dominance} & \colorbox{red!35}{0.02} \\
\footnotesize  \colorbox{red!30}{\<البند 3 (أ) من جدول الأعمال>}	 & \colorbox{red!30}{Draft resolution submitted by Argentina} & \colorbox{red!30}{0.11 } \\
\footnotesize  \colorbox{red!20}{\<ألف - آراء بشأن تنظيم الدورة>}	 & \colorbox{red!20}{plenary meetings have been made.} & \colorbox{red!20}{0.27 } \\
\footnotesize  \colorbox{red!15}{\<نفس الوطن .>}	 & \colorbox{red!15}{the same home, yes! we are brothers!} & \colorbox{red!15}{0.43 } \\
\footnotesize  \colorbox{red!10}{\<خامساً- اعتماد التقرير>}	 & \colorbox{green!5}{Adoption of the report} & \colorbox{green!5}{0.52}\\  \hdashline
\footnotesize  \colorbox{green!12}{\<ألبانيا والكاميرون.>}	 & \colorbox{green!15}{Albania, Cameroon.} & \colorbox{green!15}{0.70 } \\
\footnotesize  \colorbox{green!18}{\<مشاورات غير رسمية>}	 & \colorbox{green!10}{Informal consultation} & \colorbox{green!10}{0.72} \\
\footnotesize  \colorbox{green!20}{\<نوارة نجم تقع في حب أحمد مكي:>}	 & \colorbox{green!20}{Nawara Negm is falling in love with Mekki}& \colorbox{green!20}{0.74} \\
\footnotesize  \colorbox{green!25}{\< نتيجة التصويت كما يلي:>}	 & \colorbox{green!25}{The voting was as follows:} & \colorbox{green!25}{0.77} \\
\footnotesize  \colorbox{green!35}{\<المجموعة العربية>}	 & \colorbox{green!35}{Arab Group} & \colorbox{green!35}{  0.91   } \\ 
\footnotesize  \colorbox{green!45}{\<وتشمل هذه التكاليف ،  تكاليف الصيانة  >}	 & \colorbox{green!45}{These costs include, maintenance} & \colorbox{green!45}{0.94} \\
         \toprule        
\end{tabular}}
\caption{A sample of MSA-English parallel sentences extracted from the Open Parallel Corpus (OPUS). We report semantic similarity  on each pair of sentences using the multilingual sentence transformer model SBERT.\colorbox{green!35}{\textbf{Green:}} Selected sentences. \colorbox{red!35}{\textbf{Red:}} Ignored sentences.}

\label{tab:_sem_sim}
\end{table*}

\noindent\textbf{United Nations Parallel Corpus.}~\newcite{ziemski2016united} introduce  a manually translated united nations (UN) documents corpus covering  the six official UN languages: \textit{Arabic, Chinese, English, French, Russian,} and \textit{Spanish}. The corpus consists of development and test sets only, each of which comprise $4$K sentences that are one-to-one alignments across all official languages. 

 \noindent\textbf{IWSLT Corpus.} Several  Arabic to English   parallel datasets were released  during IWSLT\footnote{\url{https://wit3.fbk.eu}.} evaluation campaigns. These include IWSLT 2012~\cite{federico2012overview}, IWSLT 2013~\cite{cettolo2013report}, IWSLT 2016~\cite{cettolo2016iwslt}, and IWSLT 2017~\cite{cettolo2017overview}.

\noindent \textbf{Arab-Acquis.}~\newcite{habash-2017-eacl} propose Arab-Acquis. It consists of $12$k  English and French sentences extracted from the JRC-Acquis corpus~\cite{steinberger2006jrc}. The foreign sentences are translated into Arabic by two professional translators. JRC-Acquis  is  a publicly available  parallel collection of legislative text of the European Union and is written in the $22$ official European languages.

\noindent \textbf{MSA MADAR Corpus.} Proposed by ~\newcite{bouamor2018madar}, this dataset. It consists of $10$k  MSA-English  evaluation sentences   manually translated using the Crowdsourcing platform~\textit{crowdFlower.com}.\footnote{\url{http://www.crowdflower.com/}.} 


\subsection{Dialectal MT Resources}
\label{subsec:MT_corpus_DIA}
There are also a number of available dialectal datasets that can be used to extend TURJUMAN. We also briefly review these here.

\noindent \textbf{APT Corpus.}~\newcite{zbib2012machine} present an Arabic-English dataset\footnote{\url{https://catalog.ldc.upenn.edu/LDC2012T09}.} covering MSA and two other Arabic dialects. It comprises $8.11$M  MSA-English sentences, $138$k Levantine-English sentences, and $38$k Egyptian-English sentences. The dataset was collected from Arabic weblogs and  translation was carried out through Amazon Mechanical Turk.\footnote{\url{https://www.mturk.com}.}

\noindent \textbf{Qatari-English Speech Corpus.} This parallel corpus comprises $14.7$k Qatari-English sentences collected by~\newcite{elmahdy2014development} from talk-show programs and Qatari TV series and translated into
English.  

\noindent \textbf{Multi-dialectal Parallel  Corpus (MDPC).}  ~\newcite{bouamor2014multidialectal} construct MDPC by selecting  $2$k  Egyptian-English sentences from the  APT corpus~\cite{zbib2012machine}. Then, native speakers from Palestine, Syria, Jordan, and Tunisia were asked to translate the sentences into their respective native dialects.  

\noindent \textbf{Parallel Arabic Dialect Corpus (PADIC).}~\newcite{meftouh-etal-2015-machine} offers PADIC, a multi-dialect corpus including MSA, Algerian, Tunisian, Palestinian, and Syrian.  PADIC consists of $6.4$K parallel sentences between MSA and all the listed dialects.

\noindent \textbf{Dial2MSA.}~\newcite{mubarak2018dial2msa} release  this parallel dialectal Arabic corpus for converting dialectal Arabic to MSA. The dataset has $6$K  tweets from four Arabic dialects: Egyptian, Levantine , Gulf ,and Maghrebi. Each of the dialects is translated into MSA by native speakers of each dialect.

  \noindent \textbf{DIA MADAR Corpus.}~\newcite{bouamor2018madar} introduce this commissioned corpus. Arabic native speakers from $25$ Arabic cities were tasked to translate $2$k English sentences each into their own native dialect.  The sentences are selected from the Basic Traveling Expression Corpus~\cite{takezawa2007multilingual}. We now review systems for Arabic MT.
  

\subsection{Arabic MT Systems}
\vspace{3mm}
\textbf{MSA MT}. Arabic MT went through different stages, including rule-based systems~\cite{bakr2008hybrid,mohamed2012transforming,salloum2013dialectal} and statistical MT~\cite{habash2009improving,salloum2011dialectal,ghoneim2013multiword}. There has been work on Arabic MT employing neural methods. For example,~\newcite{almahairi2016first}  propose an Arabic $\leftrightarrow$ English NMT using a vanilla attention-based NMT model of \newcite{bahdanau2014neural}. Also,~\newcite{junczys2016neural} report an experimental study where phrase-based NMT across $30$ translation directions including Arabic is investigated. Other sentence-based Arabic $\leftrightarrow$ English NMT systems training on various datasets are presented in~\newcite{akeel2014ann},~\newcite{durrani2017qcri}, and~\newcite{alrajeh2018recipe}. A number of Arabic-related NMTs were also proposed to translate from languages other than English into Arabic. This includes from Chinese~\cite{aqlan2019arabic}, Turkish~\cite{el2019translating},  Japanese ~\cite{noll2019simple}, and four foreign languages\footnote{English, French, German, and Russian.} into MSA~\cite{nagoudi2022_arat5}.\\

\textbf{Dialectal MT}.  Some work has  focused on translating between MSA and Arabic dialects. For instance, \newcite{zbib2012machine} show the  impact  of  combined    MSA and dialectal  data  on dialect/MSA $\rightarrow$ English MT  performance. \newcite{sajjad2013translating} use MSA as a pivot language for translating Arabic dialects into English.~\newcite{salloum2014sentence} investigate the effect of sentence-level dialect identification and several linguistic features for dialect/MSA $\rightarrow$  English translation. \newcite{guellil2017neural} propose an NMT system for Arabic dialects using a vanilla recurrent neural network encoder-decoder model for translating Algerian Arabic written in a mixture of Arabizi and Arabic characters into MSA. \newcite{baniata2018neural} present an NMT system to translate Levantine (Jordanian, Syrian, and Palestinian) and Maghrebi (Algerian, Moroccan, and Tunisia) into  MSA. \newcite{sajjad2020arabench} introduce AraBench, an evaluation benchmark for dialectal Arabic to English MT and several NMT systems using several training settings: fine-tuning, data augmentation, and back-translation.~\newcite{farhan2020unsupervised} propose  an unsupervised dialectal NMT where the source dialect is not represented in training data (i.e., zero-shot MT \cite{lample2018phrase}). More recently,  \newcite{nagoudi-2021-Code-Mixed} introduce a transformer-based MT system for translating from code-mixed Modern Standard Arabic and Egyptian Arabic into English.~\newcite{nagoudi2022_arat5} Finally, propose three Arabic text-to-text transformer (AraT5) models dedicated to MSA and a diverse set of Arabic dialects. The models are used in several  dialects~$\rightarrow$ English MT  tasks. To the best of our knowledge, neither MSA nor dialectal machine translation systems described in this section have been made publicly available for research. 
\subsection{Open Source Arabic Tools}

There have been many efforts to develop tools to support Arabic NLP. Some tools target morphosyntax such as in  morphological analysis, disambiguation, POS tagging, and diacritization~\cite{2014-madamira,2016-farasa,camel2020}, while others focus on social meaning tasks such as sentiment analysis, emotion, age, gender, and sarcasm detection~\cite{farha2019mazajak,abdul2019aranet}. For MT, we do not know of any publicly available Arabic MT tools (let alone ones that afford many-to-Arabic translations nor diverse decoding). We now review Transformer-based Arabic and multilingual encoder-decoder models since these can be fine-tuned for MT.

\subsection{Pre-Trained Language  Models}\label{subsec:LM}

\noindent\textbf{mBART50}~\cite{liu2020multilingual} is a multilingual encoder-decoder model primarily intended for MT. It is pre-trained by denoising full texts in $50$ languages, including Arabic. Then, mBART is fine-tuned on parallel MT data under three settings: many-to-English, English-to-many, and many-to-many. The parallel MT data used contains a total of $230$M parallel sentences and covers high-, mid-, and low-resource languages.

\noindent\textbf{mT5}~\cite{xue2020mt5} is the multilingual version of \textbf{T}ext-\textbf{t}o-\textbf{T}ext \textbf{T}ransfer \textbf{T}ransformer model (T5)~\cite{raffel2019exploring}. The basic idea behind  this  model is to  treat every  text-based language task as a ``text-to-text" problem, (i.e. taking text format as input and producing new text format as output), where a multi-task learning set-up is applied to several NLP tasks: question answering, document summarization, and MT.  The mT5 model is pre-trained on the ``mC4: Multilingual Colossal Clean Crawled Corpus", which is  $\sim26.76$TB for $101$ languages (including Arabic). 

\noindent\textbf{AraT5}~\cite{nagoudi2022_arat5} is an  Arabic text-to-text Transformer  model dedicated to MSA and Arabic dialects. Again, AraT5 is an encoder-decoder Transformer similar in configuration and size to T5~\cite{raffel2019exploring}. AraT5 was trained on more than $248$GB of Arabic text ($70$GB MSA and $178$GB tweets).  In addition to Arabic, AraT5's vocabulary covers $11$ others languages. Namely, the model covers vocabulary from Bulgarian, Czech, English, French, German, Greek, Italian, Portuguese, Russian, Spanish, and Turkish. 

\begin{table}[h]
\footnotesize 
\centering
 \renewcommand{\arraystretch}{1}
\begin{tabular}{lccHccc}
\toprule
\textbf{xx$\rightarrow$ar}            & \textbf{Orig. OPUS} & \textbf{Filtering} & \textbf{Filtered-data} & \textbf{Train} & \textbf{Dev} & \textbf{Test} \\
\toprule
\textbf{bg}  & $2$M                  & sim               & $1.12$M                   & $1$M                  & $2$K              & $2$K               \\
\textbf{bs}  & $2$M                  & rand            & $1.00$M                   & $1$M                  & $2$K              & $2$K               \\
\textbf{cs}  & $2$M                  & sim               & $1.11$M                   & $1$M                  & $2$K              & $2$K               \\
\textbf{da}  & $2$M                  & sim               & $0.93$M                  & $0.93$M              & $2$K              & $2$K               \\
\textbf{de}  & $2$M                  & sim               & $0.99$M                  & $0.99$M              & $2$K              & $2$K               \\
\textbf{el}  & $2$M                  & sim               & $1.13$M                   & $1$M                  & $2$K              & $2$K               \\
\textbf{en}  & $2$M                  & sim               & $1.53$M                   & $1$M                  & $2$K              & $2$K               \\
\textbf{es}  & $2$M                  & sim               & $1.47$M                   & $1$M                  & $2$K              & $2$K               \\
\textbf{fr}  & $2$M                  & sim               & $1.17$M                   & $1$M                  & $2$K              & $2$K                \\
\textbf{hi}  & $2$M                  & sim               & $0.81$M                  & $0.81$M              & $2$K              & $2$K               \\
\textbf{it}  & $2$M                  & sim               & $1.10$M                   & $1$M                  & $2$K              & $2$K               \\
\textbf{ko}  & $2$M                  & sim               & $0.83$M                & $0.83$M              & $2$K              & $2$K               \\
\textbf{pl}  & $2$M                  & rand            & $1.00$M                   & $1$M                  & $2$K              & $2$K               \\
\textbf{pt}  & $2$M                  & sim               & $1.05$M                   & $1$M                  & $2$K              & $2$K               \\
\textbf{ru}  & $2$M                  & sim               & $1.34$M                   & $1$M                  & $2$K              & $2$K               \\
\textbf{tr}  & $2$M                  & sim               & $2$M                      & $1$M                  & $2$K              & $2$K               \\ \hdashline
\textbf{ceb} & $83.1$K               & all               & $83.1$K                   & $82.1$K               & $200$               & $200$                \\ 
\textbf{gd}  & $19.9$K               & all               & $19.9$K                   & $19.5$K               & $200$               & $200$                \\
\textbf{tmh} & $2.6$K                & all               & $2.6$K                    & $2.8$K                & $100$               & $100$                \\
\textbf{yo}  & $1.4$K                & all               & $1.4$K                    & 1.$2$K                & $100$               & $100$                \\ 
\toprule
\end{tabular}
\caption{OPUS filtering process and  data distribution in AraOPUS-20. Filtering methods: \textbf{(1) \textit{sim:}} keep only $1$M sentences with semantic similarity between {[}$0.7$,$0.99${]} \textbf{\textit{(2) random:}} if the language is not supported by SBERT model, we pick a random $1$M  pair of sentences. \textbf{\textit{(3) all}}: for the low resource languages, we keep all the parallel data.  \textbf{ar}: Arabic \textbf{bg}: Bulgarian. \textbf{bs}: Bosnian. \textbf{cs}: Czech. \textbf{da}: Danish. \textbf{de}: German. \textbf{el}: Greek.  \textbf{en}: English. \textbf{es}: Spanish. \textbf{fr}: French. \textbf{hi}: Hindi. \textbf{it}: Italian. \textbf{ko}: Korean. \textbf{pl}: Polish. \textbf{pt}: Portuguese. \textbf{ru}: Russian.  \textbf{tu}: Turkish.  \textbf{ceb}: Cebuano. \textbf{gd}: Scots Gaelic. \textbf{tmh}: Tamashek.  \textbf{yo}: Yoruba.} \label{tab:dist_opus} 
\end{table}





\section{AraOPUS-20 Parallel Dataset}\label{sec:AraOPUS-20-Data}

In this section, we describe AraOPUS-20 (the dataset we use to develop TURJUMAN) and the cleaning process we employ to ensure high quality of the data.
\newline

\subsection{Training Data.}  As mentioned earlier, in order to develop  our TURJUMAN  tool, we  use AraOPUS-20. AraOPUS-20 is extracted from OPUS~\cite{OPUS} as follows:

\begin{enumerate}
    \item  We randomly pick $2$M Arabic parallel sentences from the $16$ highest-resource languages from among our $20$ languages. Namely, we extract parallel data involving Arabic (mainly MSA) and Bulgarian, Czech, English,  French, German, Greek, Italian, Portuguese, Russian, Spanish, Hindi, Polish, Korean, and Turkish.
    
    \item We also use available data form the four low resource languages:  Cebuano (Philippine), Scots Gaelic (Scotland), Tamashek (Mali), and Yoruba (Nigeria).\footnote{We list the country where a language is mostly spoken, otherwise a given language can be spoken in more than one country.}  
\end{enumerate}

\subsection{Quality of Parallel Data.} In order to investigate  the  quality of OPUS Arabic parallel sentences,   we measure semantic similarity between the parallel sentences by running a multilingual sentence Transformer model~\cite{reimers2020making} on each pair of sentences,\footnote{We exclude the low-resource data from our semantic similarity steps as these languages are not supported in~\newcite{reimers2020making} model.} keeping only pairs with a semantic similarity score between $0.70$ and $0.99$. This allows us to filter out sentence pairs whose source and target are identical (i.e.,  similarity score = $1$) and those that are not good translations of one another (i.e., those with a cross-lingual semantic similarity score $< 0.70$). Manually inspecting the data, we find that a threshold of $> 0.70\%$ safely guarantees acquiring semantically similar (i.e., good translations) and distinct pairs of sentences (i.e., sentences from two different languages). Table~\ref{tab:_sem_sim}  shows a sample of MSA-English parallel sentences extracted from OPUS, along with their measured semantic similarity. We pick the top $1$M sentences\footnote{For low resource languages we use all available sentences.} (i.e., sentences with high semantic similarity score and satisfy our semantic similarity condition) from each language. We then split the resulting dataset into Train, Dev, and Test (see next section) and refer to the resulting benchmark that covers $20$ languages as \textit{AraOPUS-20} as we explained. \\

\subsection{Development and Test Data.} \label{subsec:valid-AraOPUS-20}

For each of development and test split, we randomly pick $2$k sentences form AraOPUS-20 (after filtering). We do this for all of the high resource languages. Regarding the low-resources languages, if the training split has more than $15$K sentences, we randomly pick $200$ sentences each for Dev and Test. Otherwise, we consider only $100$ sentences for each of these splits per language. More details about the AraOPUS-20 parallel data distribution are given in Table~\ref{tab:dist_opus}. \\


\begin{table*}[t]
    \centering 

        \begin{tabular}{lrH}
        \toprule
\small\bf {\# Decoding } & \small \textbf{Source/Translated Sentences}~~~~~~~~~~~~~~~~~~~~~~~~~~~~~~~~~~~~~~~~~~ &    \\  \toprule

\multicolumn{2}{l}{\small \colorbox{blue!13}{\textbf{en2ar:}} {\it She sort of grew up in front of everyone in Arkansas. Then as the spokesman for President Trump}} & \small    \\ \hline
\bf \small GS & \small 
\< لقد نشأت في وسط كل شخص في أركنساس ، ثم أصبحت متحدثة باسم الرئيس ترامب.>
&  \small   \\ 
           \small  \bf BS  & \small
\<لقد نشأت أمام الجميع في أركنساس ، ثم أصبحت متحدثة باسم الرئيس ترامب.>

&\small   \\ 
 \small   \bf Top-k &\small

\<لقد نشأت في وسط كل شخص في أركنساس ، ثم أصبحت متحدثة باسم الرئيس ترامب.>
&\small   \\  
		   \small   \bf Top-p  &\small
\<ترعرعت ، بجانب كل شخص في أركنساس ، ثم أصبحت الناطق باسم الرئيس ترمب ،>
		   &\small   \\ \hline

        \multicolumn{2}{l}{\small\colorbox{red!13}{\textbf{fr2ar:}} {\it Match Algérie et Cameroun : la grande victoire des fake news.}} & \small    \\ \hline
 
         
     \small  \bf GS &\small  \<مباراة الجزائر والكاميرون : النصر الكبير للأخبار الكاذبة.>
     &  \small   \\ 
           \small  \bf BS  & \small
\<مباراة الجزائر والكاميرون : الانتصار الكبير للأخبار المزيفة.>
           &\small   \\ 
           \small  \bf Top-k & \small
        \<مباراة الجزائر والكاميرون : النصر الكبير للأخبار الكاذبة.> 
           &\small   \\ 
            \small  \bf Top-p  &  \small 
            \<مباراة الجزائر والكاميرون : البطولة الرئيسة للأخبار المزيفة.>
            
            &\small   \\ \hline

\multicolumn{2}{l}{\small\colorbox{orange!20}{\textbf{pt2ar:}} {\it Já o governo federal não explicou os motivos da manutenção dos contratos}} & \small Negative  \\ \hline
\bf \small GS & \small  
\< لم تعد الحكومة الاتحادية تفسر أسباب استمرار العقود.>
  &  \small Negative \\ 
           \small  \bf BS  & \small \<لم تعد الحكومة الاتحادية تفسر أسباب الإبقاء على العقود.>
           &\small Negative \\ 
          \small   \bf Top-k & \small
          \<لم تعد الحكومة الاتحادية تفسر أسباب استمرار العقود.>
          &\small Negative \\ 
           \small   \bf Top-p  &\small\<لم تعد الحكومة الاتحادية تشرح أسبابه المتعلقة بالمحافظة على العقود بعد الآن.>
        &\small Negative \\  
\hline

             \multicolumn{2}{l}{\colorbox{red!25}{\small\textbf{ru2ar}}: {\it Резкий скачок цен на зерновые из-за войны: что будет с ценами на продовольствие?}} & \small    \\ \hline
 \bf \small GS & \small  
\< ارتفاع أسعار الحبوب بسبب الحرب : ماذا عن أسعار الغذاء ؟>
 &  \small   \\ 
           \small  \bf BS  & \small
\< ارتفاع حاد في أسعار الحبوب بسبب الحرب : ماذا عن أسعار الغذاء ؟>
           &\small   \\ 
           \small   \bf Top-k &\small  
           \<ارتفاع أسعار الحبوب بسبب الحرب : ماذا عن أسعار الغذاء ؟>
           &\small   \\  
		   \small   \bf Top-p  &\small 
		   \<تناقص حاد في أسعار الأمونيا بسبب الحرب : ماذا لو تناقصت أسعار الغذاء ؟>
		    &\small   \\ \hline
           
           \multicolumn{2}{l}{\colorbox{green!13}{\small\textbf{tr2ar}}: {\it Türkiye ile Ermenistan arasında, son dönemde yeniden başlayan doğrudan uçuşlardan birindeyiz.}} & \small    \\ \hline
 \bf \small GS & \small 
\<نحن في رحلة طيران مباشرة بين تركيا وأرمينيا ، التي بدأت في الآونة الأخيرة.>
 &  \small   \\ 
           \small  \bf BS  & \small\<بين تركيا وأرمينيا ، نحن في واحدة من الرحلات الجوية المباشرة التي بدأت من جديد خلال الفترة الأخيرة.>
           &\small   \\ 
           \small   \bf Top-k &\small
           \<نحن في رحلة طيران مباشرة بين تركيا وأرمينيا ، التي بدأت في الآونة الأخيرة>
           &\small   \\  
		   \small   \bf Top-p  &\small 
		   \<حيث أنه يتم نقل معظم الركاب الطيارين ضمن عصر النقل الجوي الداخلي.>
		   &\small   \\ \hline
             \toprule 

        \end{tabular}
    \caption{A sample of sentences from five foreign languages along with their MSA translations using four decoding methods. \textbf{GS}: Greedy Search.   \textbf{BS}: Beam Search.  \textbf{Top-k}: top-k sampling.   \textbf{Top-p}: top-p sampling. \colorbox{blue!13}{\textbf{en2ar:}} English to Arabic.  \colorbox{red!13}{\textbf{fr2ar}} French to Arabic. \colorbox{orange!20}{\textbf{pt2ar:}} Portuguese to Arabic. \colorbox{red!25}{\textbf{ru2ar:}} Russian to Arabic. \colorbox{green!13}{\textbf{tr2ar:}} Turkish to Arabic. }
    
    \label{tab:decoding}
\end{table*}

\begin{table*}[t]
\renewcommand{\arraystretch}{1.2}
\begin{tabular}{lllc}
\toprule
\textbf{Arguments}                 & \textbf{Description}                                          & \textbf{Command-line    }              & \textbf{Required} \\
\toprule
\textit{- - help or -h}       & show the help message and exit                                   & interactive, translate        & no      \\
\textit{- - text or -t }       & translate the input text into Arabic                                    & translate        & yes      \\
\textit{{- - input\_file}}              & path of input file                                   & translate                     & yes      \\
\textit{{- - batch\_size or -bs }}            & the number of sentences translated in one
                        iteration                      &  translate & no   \\
                        \textit{{- - seq\_length or -s}}            & generate sequences of maximum length \textit{seq\_length }                      & interactive, translate & no   \\

\textit{{- - search\_method or -m}}           &   decoding   method [\textit{`greedy'}, \textit{`beam'}, \textit{`sampling'}]                                                 & interactive, translate        & no       \\
\textit{{- - n\_beam}}                 & beam serach with a size of \textit{n\_beam}                   & interactive, translate        & no       \\
\textit{{- - top\_k or -k}}           & sampling using \textit{top-k}                                  & interactive, translate        & no       \\
\textit{{- - top\_p or -p}}           & sampling using \textit{top-p}                                  & interactive, translate        & no       \\
\textit{{- - no\_repeat\_ngram\_size}} & ngram size cannot be repeated in the generation & interactive, translate & no      \\
\textit{{ - -max\_outputs or -o  }}            & number of hypotheses to output                       & interactive, translate & no   \\
\textit{{- - cache\_dir or -c}} & path of the cache directory & interactive, translate & no      \\ 
\textit{{- -logging\_file ot -l } } & the logging file path  & interactive, translate   & no      \\  
\textit{{- - hyp\_file or -p}}                     & path of hypothesis file                              & score                         & yes      \\
\textit{{- - ref\_file  or -g}}                     & path of references file                              & score                         & yes      \\
 \toprule
\end{tabular}

\caption{\textit{Required} and \textit{optional} arguments  for each of TURJUMAN command-line tools.}

\label{tab:agrs}
\end{table*}



\begin{table*}[t]
 \begin{center}
\begin{adjustbox}{width=\textwidth}
\renewcommand{\arraystretch}{1.8}
{
\begin{tabular}{lccccccccccccccccc|cccc}

\toprule
 \textbf{Model} & \textbf{Split} & { \textbf{bg}}& { \textbf{bs}}& { \textbf{cs}} & { \textbf{da}} & { \textbf{de}} & { \textbf{el}} & { \textbf{en}} & { \textbf{es}} & { \textbf{fr}} & { \textbf{hi}} & { \textbf{it}} & { \textbf{ko}} & { \textbf{pl}} & { \textbf{pt}} & { \textbf{ru}} & { \textbf{tr}}& { \textbf{ceb}} & { \textbf{gd}} & { \textbf{tmh}} & { \textbf{yo}}\\ \toprule
\textbf{\multirow{2}{*}{\textbf{S2SMT} }} & \textbf{Dev} &{5.45}& {4.14}& {4.73}& {5.04}& {4.28}& {6.17}& {7.08}& {6.42}& {7.64}& {4.45}& {5.27}& {3.85}& {6.59} & {7.24}& {5.41}& {4.27}& {1.67}& {0.13} & {0.23}& {2.59} \\
& \textbf{Test}  &{5.25}& {4.44}& {4.38}& {5.94}& {4.61}& {6.77}& {7.42}& {6.22}& {6.98}& {4.59}& {5.24}& {3.58}& {6.93} & {7.15}& {5.65}& {4.25}& {1.98}& {0.17} & {0.0.25}& {2.48} \\
\cline{1-22} 
 \textbf{\multirow{2}{*}{\textbf{mBART} }} & \textbf{Dev}& {-}& {-}& {0.98}& {-}& {0.71}& {-}& {8.02}& {1.23}& {1.45}& {0.39}& {0.22}& {0.47}& {1.18}& {1.82}& {1.93}& {2.03}& {-}& {-}& {0.02}& {-}\\
& \textbf{Test}& {-}& {-}& {1.03}& {-}& {0.88}& {-}& {8.38}& {1.36}& {1.66}& {0.40}& {0.32}& {0.39}& {1.38}& {1.86}& {1.78}& {2.15}& {-}& {-}& {0.06}& {-}\\ \cline{1-22} 

 \textbf{\multirow{2}{*}{\textbf{mT5} }} & \textbf{Dev}& {8.13}& {6.63}& {9.71}& {10.94}& {16.64}& {\underline{14.28}}& {25.41}& {21.12}& {19.04}& {\underline{4.29}}& {15.17}& {\underline{6.08}}& {\underline{3.29}}& {10.96}& {26.63}& {10.84}& {10.23}& {\underline{2.37}}& {\underline{0.58}}& {3.45} \\ 
& \textbf{Test}& {12.85}& {6.60}& {7.79}& {10.94}& {14.90}& {\textbf{15.33}}& {25.24}& {21.12}& {20.74}& {\textbf{4.80}}& {12.90}& {\textbf{8.27}}& {\textbf{6.41}}& {21.13}& {27.94}& {11.77}& {7.53}& {\textbf{3.20}}& {\textbf{0.43}}& {\textbf{5.24}}\\ \toprule

\textbf{\multirow{2}{*}{\textbf{TURJ} }}&\textbf{Dev}&{ \underline{8.68}} &{\underline{7.94}} &{\underline{9.83}} &{\underline{11.30}} &{\underline{16.84}} &{13.82} &\underline{25.80} &{\underline{21.57}} &{\underline{21.43}} &{2.86} &{\underline{16.99}} &{2.18} &{3.43} &{\underline{12.00}} &{\underline{29.67}} &{\underline{11.75}} &{\underline{9.89}} &{2.32} &{0.64} &{\underline{5.11}}\\

&\textbf{Test} & {\textbf{13.64}} &{\textbf{7.87}} &{\textbf{8.32}} &{\textbf{11.30}} &{\textbf{16.05}} &{15.06} &\textbf{25.46} &{\textbf{21.57}} &{\textbf{22.43}} &{3.29} &{\textbf{14.92}} &{3.44} &{6.38} &{\textbf{23.64}} &{\textbf{31.68}} &{\textbf{13.05}} &{\textbf{8.43}} &{2.41} &{0.29} &{{{4.39}}} \\

 \toprule

\end{tabular}}
\end{adjustbox}
\caption{Results of TURJUMAN in BLEU on Dev and Test splits of AraOPUS-20 dataset. \textbf{Bolded}: best result on Test. \underline{Underlined}: best result on Dev. \textbf{bg}: Bulgarian. \textbf{bs}: Bosnian. \textbf{cs}: Czech. \textbf{da}: Danish. \textbf{de}: German. \textbf{el}: Greek. \textbf{en}: English. \textbf{es}: Spanish. \textbf{fr}: French. \textbf{hi}: Hindi. \textbf{it}: Italian. \textbf{ko}: Korean. \textbf{pl}: Polish. \textbf{pt}: Portuguese. \textbf{ru}: Russian. \textbf{tu}: Turkish. \textbf{ceb}: Cebuano. \textbf{gd}: Scots Gaelic. \textbf{tmh}: Tamashek. \textbf{yo}: Yoruba. \textbf{Dash} (\textbf{{$-$}}):~language is not supported by mBART50 many-to-many.}
\label{tab:res-test}
 \end{center}
\end{table*}

\begin{table}[t]
\centering
 \renewcommand{\arraystretch}{1.3}
\resizebox{0.9\columnwidth}{!}{%
\begin{tabular}{lccccc}
\toprule 
\textbf{Model}& \textbf{Split} & \textbf{en} &  \textbf{es}&\textbf{fr} & \textbf{ru} \\\toprule
\textbf{\multirow{2}{*}{\textbf{S2SMT} }} & Dev & $19.79$&$17.03$	&$13.47$ 	&$15.84$		\\ 
                                          & Test & $18.66$ &	$17.56$	&$14.38$ 	&$14.61$	\\ \hline
\textbf{\multirow{2}{*}{\textbf{mBART} }} & Dev & $9.95$&$1.78$	&$2.03$	&$1.27$		\\
                                          & Test & $9.65$&$1.86$	&$2.12$	&$1.96$	\\\hline
                                          
\textbf{\multirow{2}{*}{\textbf{mT5} }} & Dev& $27.68$&$23.54$	&$20.22$	&$20.09$\\
                                          & Test & $29.93$&$25.49$	&$21.66$	&$20.94$	 \\\hline

\textbf{\multirow{2}{*}{\textbf{TURJ} }} & Dev & \underline{$30.54$}&\underline{$26.21$}	&\underline{$22.82$}	&\underline{$22.87$}	 	\\
			
                                          & Test & $\bf32.07$ &	$\bf28.16$	&$\bf24.11$ 	&$\bf23.95$	\\
                                          			
\toprule

\end{tabular}%
}
\caption{Results of TURJUMAN in BLEU on Dev and Test splits of UN dataset. \textbf{en}: English. \textbf{es}: Spanish. \textbf{fr}: French. \textbf{ru}: Russian.}
\label{tab:un-res-test}
\end{table}
\section{TURJUMAN Tool} \label{sec:Torjuman}

TURJUMAN is a publicly available toolkit for translating sentences from $20$ languages into MSA. The package consists of a Python library and related command-line scripts. In this section, we discuss: \textbf{(1)} the training and evaluation processes of TURJUMAN's backbone MT model and \textbf{(2)} how we design the TURJUMAN tool itself and its different settings.

\subsection{Approach}

\textbf{Training.} For all the $20$ languages, we fine-tune  AraT5~\cite{nagoudi2022_arat5} with training data of AraOPUS-20 (see in Section~\ref{sec:AraOPUS-20-Data}). That is, we train a single \textit{multilingual} model that translates from a given foreign language into MSA (many-to-MSA).  Currently, a user needs to specify the identity of the input language.\footnote{We plan to incorporate a language ID module into TURJUMAN in the future. } We train our models on $96$~AMD MI50 GPUs ($16$GB each) for $25$ epochs with a batch size of $32$, maximum sequence length of $256$ tokens, and a learning rate of $5e^{-5}$. \\

\textbf{Evaluation.} In order to evaluate our TURJUMAN model, we use  two  datasets (AraOPUS-20 and United Nations Parallel Corpus, both described in Section~\ref{subsec:valid-AraOPUS-20}).\footnote{We exclude the Chinese language as it is not included in our training data.} As a rule, for all datasets we identify the best model on our Dev data\footnote{We merge all the development data for all the $20$ languages.} and blind-test it on our Test split for each language separately. For the two datasets,  we report results on both Dev and Test splits as shown in Tables~\ref{tab:res-test} and~\ref{tab:un-res-test} respectively.\\ 

    
    

\textbf{Baselines.}
For comparison,  we use three baselines:

\begin{itemize}
    \item \textbf{Baseline I}. A vanilla sequence-to-sequence (S2S) Transformer~\cite{vaswani2017attention} as implemented in Fairseq~\cite{ott2019fairseq}. We train this model from scratch using AraOPUS-20 training data  
     \item \textbf{Baseline II}.  We fine-tune the multilingual encoder-decoder model mT5~\cite{xue2020mt5} on the same training data as our second baseline.
     \item \textbf{Baseline III}. We use the mBART-50 many-to-many multilingual MT model for our third baseline. We do not fine-tune  this model on AraOPUS-20 Train data as it is a checkpoint of mBART-large-50~\cite{liu2020multilingual} already fine-tuned on a multilingual MT dataset covering high-, mid-, and low-resource languages. In total, this model is fine-tuned with $230$M parallel sentences from these $50$ languages.

\end{itemize}

\subsection{Implementation}

We distribute TURJUMAN as a modular toolkit built around standard libraries including PyTorch \cite{paszke2019pytorch} and HuggingFace~\cite{lhoest2021datasets}. \\

\textbf{Command-Line Tools.} We provide several command-line tools for translation and evaluation: 

\begin{itemize}
    \item \textit{turjuman\_interactive:} This interactive command line facility can be used for quick sentence-by-sentence translation exploiting our fine-tuned NMT model (TURJUMAN's backbone translation model). 
    \item \textit{turjuman\_translate:} This is the same as the interactive command. However, a path to a file containing source sentences is required.
    \item  \textit{turjuman\_score:} This evaluates an output translation (output translations) against reference translation(s) in terms of a BLEU score.  
\end{itemize}

\subsection{Decoding Support} \label{sec:MT-decoding}

We also endow TURJUMAN with support for MT-based \textit{paraphrase} generation by adding four decoding methods at the decoder side. We implement a number of prominent decoding methods used in the literature. Namely, we implement \textit{greedy search}, \textit{beam search} \cite{koehn2009statistical}, \textit{top-k sampling}  \cite{fan2018hierarchical}, and \textit{nucleus sampling} \cite{holtzman2019curious}. Table~\ref{tab:decoding} shows example translations with TURJUMAN exploiting each of the four decoding methods. We now briefly describe each of these methods. \\

\textbf{Greedy Search.} Is  a simple heuristic strategy aiming to selects the word with the highest conditional probability as its next word at each timestep as shown in Formula~(1): 

\begin{equation}
w_t= argmax_wP(w | w_{1:t-1})
\end{equation}\label{eq:1}

\textbf{Beam Search.} Beam search is an improved version of greedy search that uses a hyper-parameter $num\_beams$. It is based on exploring the solution space and reduces the risk of missing hidden high probability word sequences by keeping the most likely $num\_beams$ of hypotheses sequences. \\

\textbf{Top-k Sampling.} A probabilistic  decoding method proposed by~\newcite{fan2018hierarchical} that aims to avoid repetitions  during decoding. This method also increase diversity of the output by using  a simple, yet  powerful sampling stochastic  scheme called \textit{\textbf{top-k}} sampling. First, the top $k$ words with the highest probability  are selected. Then, we sample from this shortlist of words. This allows the other high-scoring tokens a chance of being picked. Formula~(2)  describe top-k sampling, where $V^{(k)}$ is the top-k vocabulary.

\begin{equation}
w_t= \sum_{w \in V^{(k)}} P(w |w_{1:t-1}) 
\end{equation}\label{eq:2}

\textbf{Top-p Sampling.} Also called \textit{nucleus-sampling}, this method is proposed by ~\newcite{holtzman2019curious}. It shares the same principle as the \textit{top-k} method, and the only difference between the two is that Top-p sampling chooses from the smallest possible set of words the sum of whose probability is greater than a certain probability $p$ (i.e., threshold). This method is described in Formula~(3),  where $V^{(p)}$ is the top-p vocabulary.

\begin{equation}
w_t= \sum_{w \in V^{(p)}}P(w |w_{1:t-1}) \geq p 
\end{equation}\label{eq:3}

\subsection{TURJUMAN  Arguments.} Each of the command-line tools (i.e., \textit{turjuman-interactive, turjuman-translate,} and t\textit{urjuman-score}) support/require several arguments. Table~\ref{tab:agrs} presents the required and optional arguments for each of these TURJUMAN tools.

\subsection{Discussion}

 Results reported  in Table~\ref{tab:res-test} show that TURJUMAN achieves best BLEU score in  $13$ out of the $20$ tests splits, outperforming all our baselines: S2SMT, mBART, and mT5 with  +$8.07$, +$11.28$, and +$0.53$ BLEU points on average.  We also note that mT5 outperforms AraT5 mostly in the languages that were not included in AraT5 vocabulary. Namely, we observe this in Hindi, Polish, Korean, Scots Gaelic, Tamashek, and Yoruba (see Section~\ref{subsec:LM}). In addition, as Table~\ref{tab:un-res-test} shows, TURJUMAN outperforms \textit{all} baselines in UN-Test data in the four investigated MT tasks: English, French, Spanish, and Russian $\rightarrow$ Arabic.

\section{TURJUMAN: Getting Started} \label{sec:Torjuman_use}

\subsection{Installation}
TURJUMAN is implemented in Python and  can be installed using the pip package manager.\footnote{\textit{pip install turjuman}} It is compatible with \textit{Python 3.6} and later versions, \textit{Torch 1.8.1}, and \textit{HuggingFace Transformers 4.5.1 library}.\footnote{Installation instructions and documentation can be found at:~\url{https://github.com/UBC-NLP/turjuman}.}

\begin{lstlisting}
> pip install turjuman 
\end{lstlisting}
\subsection{Turjuman Command Line Examples}

As explained, TURJUMAN provides several command-line tools for translation and evaluation. Each command supports multiple arguments. In the following, we provide a number of examples illustrating how to use TURJUMAN command-line tools with different arguments. 
  
\begin{itemize}[leftmargin=*]
    \item \textbf{turjuman\_interactive.} In the following two examples we use \textit{turjuman\_interactive} to generate translations interactively for an English and Portuguese sentences, respectively. Here, we use beam search with a beam size of $5$, a maximum sequence length of $300$, and a number of targets to output at $3$.
\end{itemize}

\begin{lstlisting}
> turjuman_interactive
\end{lstlisting}
\textcolor{blue}{Output--------------------------------------------------------}
\begin{topbot}
\small
$>$ \texttt{Turjuman Interactive CLI} \\
$>$ \texttt{Loading model from UBC-NLP/turjuman} \\
$>$ \texttt{\textcolor{red}{Type your source text or (q) to STOP:}} \\\\
$>$ \texttt{\textbf{She thought a dark moment in her past was forgotten.}} \\
$>$ \texttt{\textcolor{green}{\textbf{target1}}: \<اعتقدت أن لحظة مظلمة في ماضيها قد نسيت.>}\\ 
$>$ \texttt{\textcolor{green}{\textbf{target2}}: \<لقد اعتقدت أن لحظة مظلمة في ماضيها قد نسيت.>}\\ 
$>$ \texttt{\textcolor{green}{\textbf{target3}}: \<كانت تعتقد أن لحظة مظلمة في ماضيها قد نسيت.>}\\\\
$>$ \texttt{\textcolor{red}{Type your source text or (q) to STOP:}} \\
$>$ \texttt{\textbf{Esta é uma lista de estados soberanos}} \\
$>$ \texttt{\textcolor{green}{\textbf{target1}}: \<هذه قائمة الدول ذات السيادة>}\\ 
$>$ \texttt{\textcolor{green}{\textbf{target2}}: \<هذه قائمة بالدول ذات السيادة>}\\ 
$>$ \texttt{\textcolor{green}{\textbf{target3}}: \<قائمة الدول ذات السيادة هذه>}\\ 
\end{topbot}

\begin{itemize}[leftmargin=*]
\item \textbf{turjuman\_translate.} In the following  we show how to use turjuman\_translate with two modes of input:
\end{itemize}     
\textbf{(1) Text.} A raw text is passed to the turjuman model directly troughs command line using the argument \textit{-text} or \textit{-t}. Translation will display directly on the terminal. \\

\begin{lstlisting}
> turjuman_translate --text "Je peux payer le traitement de votre fille"
\end{lstlisting}
\textcolor{blue}{Output--------------------------------------------------------}
\begin{topbot}
\small
$>$ \texttt{Turjuman Translate CLI} \\
$>$ \texttt{Translate from input sentence} \\
$>$ \texttt{Loading model from UBC-NLP/turjuman} \\
$>$ \texttt{\textcolor{green}{\textbf{target}}: \<يمكنني أن أدفع ثمن علاج ابنتك>} \\


\end{topbot}

\textbf{(2) File.} The argument \textit{-input\_file} or \textit{-f } can be used to import a set of sentences from a text file. Translation will be saved on a JSON file format.  \\

\begin{topbot} 
\begin{lstlisting}
> turjuman_translate --file ./sample.txt
\end{lstlisting}
\textcolor{blue}{Output-----------------------------------------------------}

\small
$>$ \texttt{Turjuman Translate CLI} \\
$>$ \texttt{Translate from samples.txt} \\
$>$ \texttt{Loading model from UBC-NLP/turjuman}\\
$>$ \texttt{\textcolor{green}{{\textbf{Translation is saved in samples.json}}}}
 
\end{topbot}








\begin{itemize}[leftmargin=*]
    \item \textbf{turjuman\_score.} This evaluates an output translation (output translations) against reference translation(s) in terms of a BLEU score. 
\end{itemize}

\begin{topbot} 
\begin{lstlisting}
> turjuman_score -p "translated_targets.txt" -g "gold_targets.txt"
\end{lstlisting}
\textcolor{blue}{Output-----------------------------------------------------}

\small
$>$ \texttt{Turjuman Score CLI} \\
$>$ \texttt{\textcolor{red}{\textbf{hyp\_file}}=translated\_targets.txt}\\
$>$ \texttt{\textcolor{blue}{\textbf{ref\_file}}=gold\_targets.txt}\\
$>$ \texttt{\textcolor{green}{{\textbf{bleu score: 43.573826221233}}}}
 
\end{topbot}

\section{Conclusion}\label{sec:concl}

We presented TURJUMAN, an open-source Python-based package and command-line tool for Arabic neural machine translation. In the context of developing TURJUMAN, we also extracted and prepared a high quality $20$ language pairs benchmark for MSA MT (\textit{AraOPUS-20}). We exploit AraOPUS-20 to fine-tune an Arabic text-to-text Transformer model, AraT5. Our resulting multilingual model outperforms competitive baselines, demonstrating the utility of our tool. In addition to its translation ability, TURJUMAN integrates a number of decoding methods. This allows for use of the tool for paraphrasing foreign sentences into diverse Arabic sequences. TURJUMAN is extensible, and we plan to train it with larger datasets and further enhance its functionality in the future. We also plan to explore adding new language pairs to TURJUMAN.

\section*{Ethical Considerations}\label{sec:ethics}
TURJUMAN is developed using publicly available data. Hence, we do not have serious concerns about personal information being retrievable from our trained model. Similar to many NLP tools, TURJUMAN can be misused. However, the tool can be deployed for a wide host of useful application such as in education or travel. We do encourage deploying TURJUMAN in socially-relevant scenarios.
\section*{Acknowledgements}\label{sec:acknow}
We gratefully acknowledge support from the Natural Sciences and Engineering Research Council of Canada (NSERC; RGPIN-2018-04267), the Social Sciences and Humanities Research Council of Canada (SSHRC; 435-2018-0576; 895-2020-1004; 895-2021-1008), Canadian Foundation for Innovation (CFI; 37771), Compute Canada (CC),\footnote{\href{https://www.computecanada.ca}{https://www.computecanada.ca}} UBC ARC-Sockeye,\footnote{\href{https://arc.ubc.ca/ubc-arc-sockeye}{https://arc.ubc.ca/ubc-arc-sockeye}} and Advanced Micro Devices, Inc. (AMD). Any opinions, conclusions or recommendations expressed in this material are those of the author(s) and do not necessarily reflect the views of NSERC, SSHRC, CFI, CC, AMD, or UBC ARC-Sockeye.



\section{Bibliographical References}\label{reference}

\bibliographystyle{lrec}
\bibliography{lrec2020W-xample-kc}

\label{lr:ref}
\bibliographystylelanguageresource{lrec}

\end{document}